\documentclass[fleqn]{llncs}
\usepackage{makeidx}  % allows for indexgeneration
\usepackage{placeins}
\usepackage{amsmath}

\makeatletter
\AtBeginDocument{%
  \expandafter\renewcommand\expandafter\subsection\expandafter
    {\expandafter\@fb@secFB\subsection}%
  \newcommand\@fb@secFB{\FloatBarrier
    \gdef\@fb@afterHHook{\@fb@topbarrier \gdef\@fb@afterHHook{}}}%
  \g@addto@macro\@afterheading{\@fb@afterHHook}%
  \gdef\@fb@afterHHook{}%
}
\makeatother

\begin{document}
\frontmatter          % for the preliminaries
\pagestyle{headings}  % switches on printing of running heads
\addtocmark{Deductive and Analogical Reasoning} % additional mark in the TOC
\mainmatter              % start of the contributions
\title{Deductive and Analogical Reasoning on a Semantically Embedded Knowledge Graph}
\titlerunning{Deductive and Analogical Reasoning}  % abbreviated title (for running head)
\author{Douglas Summers-Stay}
\authorrunning{Douglas Summers-Stay} % abbreviated author list (for running head)
\tocauthor{Douglas Summers-Stay}
\institute{U.S. Army Research Laboratory,\\
\email{douglas.a.summers-stay.civ@mail.mil}}

\maketitle              % typeset the title of the contribution

\begin{abstract}
Representing knowledge as high-dimensional vectors in a continuous semantic vector space can help overcome the brittleness and incompleteness of traditional knowledge bases. We present a method for performing deductive reasoning directly in such a vector space, combining analogy, association, and deduction in a straightforward way at each step in a chain of reasoning, drawing on knowledge from diverse sources and ontologies.
\keywords{semantic vectors, reasoning, knowledge graphs, knowledge bases, analogy}
\end{abstract}
\vspace{-25pt}
\section{Introduction}
\vspace{-10pt}
Common sense knowledge bases (KB) are notoriously `brittle': they are generally only usable by those who have spent a lot of time getting to know precisely how to phrase a question so that it will match the representation in the KB\cite{Buchanan}. They are also inevitably incomplete, leaving out many facts that one would expect a system that claims common sense to include. In order to get around these limitations, several researchers\cite{Freitas}\cite{West}\cite{Summers-Stay} have been exploring the possibililty of somehow combining the deductive reasoning abilities of a knowledge base with the ability to represent semantic similarity that is provided by distributional semantic vector spaces. ``Query expansion,'' for example, involves querying for semantically nearby terms as well as the explicit terms entered. The deductive reasoning in such a system still takes place in the discrete knowledge base, however. When there are concepts or relations missing from the knowledge base that prevent a chain of reasoning from going through for any of these near terms,the system will be unable to return any result.

Searches that take place completely in a semantic vector space, on the other hand, are more akin to searching via a web search engine. These searches forgo any explicit steps of deductive reasoning, relying instead on broad coverage. Combining multiple facts in a chain of reasoning to answer a query is beyond their current capabilities.
What we propose in this paper is a way of discovering chains of reasoning connecting a premise to a conclusion directly in a semantic vector space. The method can be applied to various ways of representing knowledge by high-dimensional vectors.

Forming a chain of deductive logical reasoning can be thought of as a special variety of a more general phenomenon in the mind of following a ``train of thought.'' One idea brings up a related idea, which in turn brings up another related idea, and so forms a connected train. We can deliberately return to an earlier point in the train and follow another path either backward or forward, so that the trains link up to form a larger structure.

Trains of thought serve several purposes. Parts of an essay or a story are often structured as trains of thought, with each sentence building on the one before. Restricted to cause-effect relations, the root cause of an event can be found. Trains formed of links between means and ends can form a plan of actions and subgoals to achieve a larger goal. Trains of reasons can answer ``why'' questions. Trains of looser relations like resemblance of form and sound form the basis of some kinds of poetry, symbolism, or mysticism.  Memory techniques, creativity methods, and dreams also rely on trains of thought.

In order to form chains of reasoning, AI researchers have attempted to find paths between ideas using exhaustive search in a knowledge graph. This blind walk through all connections in the graph seems very different from how we normally think. A path connecting two ideas seems to bubble up-- we initially feel the connection more than see it.  Ideas shade imperceptibly into one another. Analogy and association are everpresent. An argument as originally conceived generally skips steps, and may include steps which are simply analogous to related problems. Turning such a jumble of ideas into a step-by-step proof is a process that takes skill, training, and deliberate conscious effort.

Such imprecision can lead to invalid conclusions and fuzzy thinking, but it has the advantage of being capable of operating under unknown or incompletely represented conditions. When we don't know, we can guess at the general ballpark of the answer. In order to create a system that can deal with the ambiguities of natural language and take action in an uncertain environment, we need to build in the ability to think in a more flexible, human manner. A more human-like reasoning engine should have at least the following properties:
\small
\begin{itemize}
\item	Be capable of associational, analogical, inductive, abductive, and deductive reasoning;
\item	when exact answers can't be found, guess at an approximate answer;
\item	be aware of the strength or weakness of its arguments;
\item	creatively find connections that were not deliberately given, and
\item	find arguments that add up to a whole, rather than find strictly linear connections.
\end{itemize}
\normalsize
\vspace{-8pt}
\section{Background}
\vspace{-5pt}
There are multiple strands of research that involve representing knowledge as vectors. One strand comes from the biologically-inspired cognitive architecture community. This is increasingly known as Vector Symbolic Architectures (VSA)\cite{Gayler}. \cite{Kanerva} introduced the idea of using sparse high-dimensional binary vectors as a way of storing information that was resistant to noise and capable of addressing memory with exemplars. These ideas have been developed to include the notion of binding vectors for compositional structure and to be more biologically accurate\cite{Widdows} \cite{Hummel} \cite{Levy}.

A second strand comes from the linguistics community, beginning with Latent Semantic Analysis to create word and document context vectors\cite{Dumais}, and includes the well-known word vector representation word2vec\cite{Mikolov}. The ability of such vectors to solve analogy problems was demonstrated in 2005\cite{Turney}. Attempts to encode the meaning of sentences by composing the meaning of the words in the sentence \cite{Kiros} \cite{Grefenstette} \cite{Baroni} is a very similar problem to encoding triples from a knowledge base. Some researchers encode triples from a knowledge graph directly as vectors, building on \cite{Bordes}. 

A few papers are directly concerned with multi-step deductive reasoning in vector spaces \cite{Lee} \cite{Widdows2}  \cite{Rocktäschel} \cite{Wang}. These approaches use machine learning to build methods for composing vectors in a reasoning chain. The system described in this paper does not require any training beyond what is done to create the word vector representations in the first place. It is unique in using sparse vector decomposition to solve a deductive reasoning problem.
\vspace{-5pt}
\section{Method}
\vspace{-5pt}
We are given a knowledge base of facts represented as triples of the form $(e_n$,$predicate$, $e_m)$. We are also given a semantic vector space where every entity $e$  is represented by a high-dimensional vector in such a way that terms that are semantically similar are nearby in the semantic space. Each of the triples is represented within the vector space by a vector of the form $ -\vec{e_1} + \vec{e_2} $. For the purposes of the vector space calculations, these triples are treated as statements that $e_n\Rightarrow e_m$. The specific predicates are not used in the vector space calculation, but instead all predicates are treated as a simple statement of implication. This maps the first-order predicate calculus problem to a ``zeroth-order'' propositional calculus problem.

We wish to prove that $g\Rightarrow p$. The vector representing this relation is $ -\vec{g} + \vec{p} $. If there is some set of facts in the knowledge base that can prove this, it must be the case that the facts form a chain:
\begin{eqnarray*} 
g\Rightarrow e_1 \Rightarrow e_2 ... \Rightarrow e_n \Rightarrow p
\end{eqnarray*}

Representing this chain as vectors we get
\begin{eqnarray*}
(-\vec{g} + \vec{e_1}) + (-\vec{e_1} + \vec{e_2}) + ... (-\vec{e_{n-1}} + \vec{e_n}) + (-\vec{e_n} + \vec{p})
\end{eqnarray*}
Cancelling out we see that this sum is equal to the vector directly from g to p:
\begin{eqnarray*}
(-\vec{g} + \vec{e_1}) + (-\vec{e_1} + \vec{e_2}) + ... (-\vec{e_{n-1}} + \vec{e_n}) + (-\vec{e_n} + \vec{p}) =   -\vec{g} + \vec{p} 
\end{eqnarray*}
Our goal, then, in order to find a chain of entities linking g to p, is to find a sum of fact vectors of the form $ (\vec{-e_m} + \vec{e_n}) $ that adds up to $ (-\vec{g} + \vec{p}) $. Such a sum can be thought of as a weight vector $\vec{w}$ multiplied by the list of fact vectors, with a weight of 1 for each fact vector included in the chain, and a weight of zero for each fact vector not included. Clearly $\vec{w}$ will be a sparse vector, with many more zeros than ones. This suggests that in order to find such a sum, we can use sparse approximation techniques such as OMP or LASSO to obtain the sparse weight vector $\vec{w}$. 

In cases where such a chain exists, this method should (when the sparse approximation is successful) return a set of facts that constitute the chain. When the chain does not exist, however, the method will return an approximation of the correct links in the path. Because the vectors come from a semantic vector space, such approximations will amount to undefined relations between closely related entities. Such gaps can be considered a kind of associational reasoning.

For example, suppose we want to find a path of relations between $G:Michael Jackson$ and $P:music$. The knowledge base contains, among many others,  the following two facts:

(Michael Jackson, is a, songwriter) and (musician, composes, music)

The proposed method would return $Michael Jackson\Rightarrow songwriter$ and $musician\Rightarrow music$, even though they don't strictly form a chain of reasoning, because $songwriter$ and $musician$ are nearby in the semantic space, and so the error in the sum is fairly small. \footnote{In some special cases, the error in one gap of the chain will largely cancel out with the error at another gap. When this happens, the system has found an analogous relation. This is discussed in the section Analogical Properties of Semantic Spaces below.} This is the core idea we hope to communicate in this paper: that sparse solvers can be used to find deductive chains in a semantic vector space, in a way that allows for analogical and associational connections where appropriate.
\vspace{-5pt}
\section{Propositional Calculus and the Logic of Subsets}
\vspace{-25pt}
\begin{table}
\caption{Loosely speaking, terms near $a+b$ will come from the set of terms near $a$ OR $b$, while terms near $a-b$ come from terms near $a$ and NOT near $B$. Here bold terms are among the eight nearest terms to ``classical'' and italic terms are those near to ``music''. The set of terms that belong to $a$ AND $b$ is a subset of $a$ OR $b$ and these terms will show up especially high in the list of terms near $a+b$. }

\begin{center}
\small{
\bgroup
\def\arraystretch{1.5}%  1 is the default, change whatever you need
\setlength{\tabcolsep}{0.3em} % for the horizontal padding
\begin{tabular}{|r|p{8.7cm}| }
\hline
  near ``classical'' & classical, \textit{classical music}, Classical, classical repertoire, Hindustani classical, contemporary, Mohiniattam, sacred choral \\
  near ``music'' & music, \textbf{classical music}, jazz, Music, songs, musicians, tunes \\
  near ``music - classical'' & \textit{music}, Rhapsody subscription, ringtone, MP3s, Polow, Napster, entertainment, \textit{Music}, \textit{tunes} \\
  near ``music + classical'' & \textbf{classical}, \textbf{\textit{music}}, \textbf{\textit{classical music}}, \textbf{jazz}, classical repertoire, \textbf{Hindustani classical}, \textbf{sacred choral}, classical guitar\\
\hline
\end{tabular}
\egroup
}
\end{center}

\end{table}
\normalsize
\vspace{-25pt}
 The system is able to perform deductive logic because it is approximately implementing propositional calculus as a logic of subsets.\footnote{Boole and DeMorgan originally formulated propositional logic as a special case of the logic of subsets. \cite{Ellerman}}  Call the universe $U$ the set of of all entities $u$ in the semantic vector space. The nearest neighbors of any entity $p$ form a subset $P$ of $U$. (These are the terms which are semantically near to $p$.) In a high dimensional semantic vector space, if a vector is a nearest neighbor of vector $\vec{a}$ or $\vec{b}$ it will also usually be a nearest neighbor of vector $\vec{a+b}$. \footnote{If $\vec{a}$ and $\vec{b}$ are approximately orthogonal unit vectors, then the similarity between the two will be $\frac{\sqrt{2}}{2}$. This is much higher than the expected similarity between any two terms selected from the space. See \cite{Widdows} for details.} This means that we can treat $+$ as the union operator: The elements of of $A \cup B$ will be the near neighbors of the vector $\vec{a+b}$. In propositional calculus, this is the OR operator, $\lor$.

The vectors in $U$ near $-\vec{a}$ are the vectors which are not near to $\vec{a}$. So $-$ can be treated as a the set complement operator $\mathsf{c}$. In propositional calculus, this is the NOT operator, $\neg$.

 In propositional calculus, $A$ implies $B$ ($A\Rightarrow B$) means that either $B$ is true, or $A$ is not true, so it can be rewritten as (NOT A) OR B. In the subset logic, this is $A \mathsf{c} \cup B$. In the vector space, then, $A\Rightarrow B$ can be represented as $-\vec{a} + \vec{b}$. 

In propositional calculus, the \textit{modus ponens} rule allows us to conclude $B$ from the two facts $A$ and $A\Rightarrow B$. In the vector space, $\vec{a}$ and $-\vec{a} + \vec{b}$ cancel to give $\vec{b}$. In a chain of implication $ A\Rightarrow B \Rightarrow C \Rightarrow D$ all the interior terms cancel, allowing us to conclude that $A \Rightarrow D$. Similarly in the vector space, the vectors $(-\vec{a} + \vec{b}) + (-\vec{b} + \vec{c}) + (-\vec{c} + \vec{d})$ simplify to the vector $-\vec{a} + \vec{d}$.\footnote{Notice that addition is used as AND rather than OR when combining $B$ with $A$ and $A\Rightarrow B$ (see the caption of Table 1 for why this is acceptable). At any rate, the notion of cancelling out with \textit{modus ponens} still holds. } In this way, the system is able to carry out \textit{modus ponens} deductive reasoning within the semantic vector space.

Propositional calculus is less powerful than predicate calculus. In order to prove that $(p,relation,q)$ one must have, in addition to the triples in the knowledge base, Horn clauses which have $(p,relation,q)$ as the conclusion (i.e. the non-negative literal). If the facts in the knowledge base passed to the solver are limited to those which have relations that participate in such Horn clauses, the chains of implication will tend to be more reasonable. In general, using this system as it currently stands requires restricting which predicates are allowed to participate in a solution. Instead of representing $snow\Rightarrow white$, we could represent the more informative statement $(madeOf, snow)\Rightarrow (hasColor, white)$. Doing this requires using vectors that bind multiple concepts to roles, as in VSA.  It is not yet clear how well the analogical or associational properties described below would work in such an architecture, however: it depends on the details of how binding is performed. 
\vspace{-5pt}
\section{Analogical and Abductive Reasoning}
\vspace{-5pt}
The ability of distributional semantic vectors such as word2vec to find analogies is not peculiar to how such vectors are trained, but should be an expected property of any system that maps semantically similar concepts to similar high-dimensional vectors. Suppose we are given the following analogy to solve: \textit{bear:hiker::shark:X}. To make it simpler, consider contexts representing the ideas \textit{woods}, \textit{sea}, \textit{predator} and \textit{tourist}, and treat any other contexts as noise. The vector for \textit{bear}, for instance, is some weighted average of (the mean of all vectors related to woods), and (the mean of all vectors related to predators) plus some noise. Thus we can rewrite the analogy as $woods + predator : woods + tourist :: sea + predator : X$.

The vector between \textit{bear} and \textit{hiker} is $- predator+ tourist+ noise$. This is very close to the vector from \textit{shark} to \textit{snorkeler}. These two vectors are so similar because the relations between the two pairs of words being connected are so similar. Since the system looks for any vector that will make the sum have as low error as possible, it could choose the relation vector between \textit{bear} and \textit{hiker} to connect the concept \textit{shark} to the concept \textit{snorkeler}: the system can make use of analogical relations to complete a chain of argument.\footnote{When a direct chain of reasoning is possible, such links won't happen-- the analogy, being inexact, has a higher cost than the direct link.} This makes it better at handling incompleteness in the knowledge base and makes it more like human reasoning, where newly encountered concepts do not need to be exact matches to those in our memories in order for us to reason about them. In everyday thinking, analogy, association and abduction are frequently used together with deduction. 

While it is possible to use the raw distributional vectors for terms themselves as entities in the vector space, we can also define other vectors in this space. The fact that the terms in a natural category like \textit{mammal} tend to already be clustered in the semantic space means that the number of such terms that can be averaged into a category vector is somewhat larger than the results in Experiment 1. We could also make use of the analogical properties of the semantic vector space to place other concepts that don't appear in the corpora, if we know some of their attributes. These techniques are useful when attempting to embed a knowledge base into the semantic vector space, where the concepts in the knowledge base may not be named by a specific English word.\footnote{Along the same lines, \cite{Yu} describes a more intricate method of locating particular word senses in the vector space.}
\vspace{-5pt}
\section{Ontology Merging}
\vspace{-5pt}
One of the major benefits of using an embedded deduction mechanism is that it simplifies the process of merging ontologies. If we are able to map both ontologies into the semantic vector space, then even if the same concept isn't mapped to the exact same term, it will be mapped to a nearby term which may be good enough for the chain of reasoning to be found. For example, suppose one ontology contained the statement (bears, eat, grubs) and another contained the statement (insects, live in, dead trees). Neither ontology defines the relation of grubs to insects, but the system would be able to make the connection between bears and dead trees (answering the question ``Why is the bear digging in a dead tree?'' for example) because of the semantic similarity of \textit{grub} to \textit{insect}.  Such a method would be especially useful when the ontology has not been hand built. Information extraction methods that extract triples from natural language sources, such as ReVerb, can be used to add facts to the knowledge base, without worrying too much about whether the entities to which triples refer are all expressed in the same way.
\vspace{-5pt}
\section{Answering Questions}
\vspace{-5pt}
The system as described so far has been finding a chain of reasoning connecting between two terms: one ``given'', and one ``to prove.''\footnote{Deductive reasoning systems typically use either forwards or backwards inference. This system uses "middle out" inference, that doesn't begin at either end but is a holistic procedure happening all along the chain at once.} However, a knowledge base is usually used with one or more variables, to find multiple possible chains that answer a query. If the possible answers can be limited to a smaller set, this system can also be used in this way, by having the ``to prove'' vector be a sum of all of the possible answers. For example, the knowledge base contains the following statements:

\small
(apple, hasColor, red), (apple, hasColor, yellow),  (apple, hasColor, green)
\normalsize

and we want to know what colors apples have. We could put in $-apple + (red + orange + yellow + green + blue + purple)$ as the query, and the result picks out these three statements as highly relevant:

\small
1.00 (apple, hasColor, red)

0.99 (apple, hasColor, green)

0.72 (apple, hasColor, yellow)

0.08 (cordon bleu, derivedFrom, blue)\footnote{Notice that the fourth, less relevant, fact is also relating a food to a color.}

\normalsize

Notice that the goal vector is a ``category vector'' as decribed in section 7. Another way to get a particular type of result is by limiting the type of relations that are in the portion of the database that is searched. For example, if one wanted to know how B was caused, the search could be limited to those facts in the database related by causal predicates, such as \textit{causes}, \textit{turns into}, \textit{has side effect}, and so forth. One way to do this, if the Horn clauses are known, is to find all relations which participate in a Horn clause that resolves to A causes B. 
\vspace{-10pt}
\section{Ordering the Chain}
\vspace{-5pt}
The results of the sparse vector decomposition define which triples might participate in the chain, but they are unordered.\footnote{In fact, they may form a multistranded rope rather than a chain-- the ``elastic-net''\cite{Zou} parameter in LASSO can be used to encourage or discourage finding alternative equally good paths for part or all of the chain.} To arrange them in order, we use the following method. All entities that participate in a triple returned by the solver, as well as the input terms, are added to a complete directed graph. Edges corresponding to relations returned by the solver are given very low weights, while edges not included are weighted based on their distance in the semantic space. Then we find the least costly path from the head input term to the tail. \footnote{A slightly more complicated cost function can be used to encourage the lowest cost path to follow analogical connections as well.} Although the system is capable of coming up with tree-like proofs to multiple entities connected by OR, we haven't yet implemented a method for finding least-cost trees.
\vspace{-8pt}
\section{Experiments}
\vspace{-5pt}
LASSO, OMP and other sparse solvers are not guaranteed to find the optimal solution (which would be an NP-complete problem.) Their performance depends on the size, dimensionality, and clustering of the data.  We characterized how well LASSO performed for the vectors in our dataset. For all these experiments, we used the 300-dimensional word2vec vectors provided by Mikolov\cite{Mikolov}. We used L=20, and lambda=.2 for the LASSO parameters.
\vspace{-5pt}
\subsection{Experiment 1}
\begin{table}
\vspace{-15pt}
\caption{How frequently all terms in sum are among 20 nearest neighbors of sum / how frequently all terms are within results of LASSO with L=20}
\begin{center}
\small{
\bgroup
\def\arraystretch{1.5}%  1 is the default, change whatever you need
\setlength{\tabcolsep}{0.3em} % for the horizontal padding
\begin{tabular}{|r|c|c|c|c|c|c|c|c|c|}
	\hline
	&	2&	3&	4&	5&	6&	7&	8&	9&	10\\
	\hline
1000&			100/100&	100/100&	97/99&	85/98&	47/98&	21/96&	5/88&	1/76&	0/50\\
10000&		100/100&	98/100&	76/100&	25/100&	3/100&	0/99&	0/98&	0/94&	0/83\\
100000&		100/100&	91/100&	45/100&	6/100&	0/97&	0/83&	0/67&	0/39&	0/11\\
1000000&		100/100&	84/97&	27/88&	2/52&	0/14&	1/1&	0/0&	0/0&	 0/0\\
\hline
Dictionary size &
\multicolumn{9}{|l|}{}\\
\hline
\end{tabular}
\egroup
}
\end{center}
\vspace{-25pt}
\end{table}	

As noted in the section on propositional calculus, it is a curious property of high-dimensional vector spaces that the vector $\vec{a+b}$ will tend to be closer to $\vec{a}$ and $\vec{b}$  than other vectors in the space, assuming they are fairly well distributed. However, this property only holds for a few vectors being added together. In Table 1, we added from 1 to 10 randomly chosen term vectors, and found how frequently all of the summed vectors were present among the 20 nearest neighbors of the sum vector, for various dictionary sizes. For larger dictionaries, fewer of the summed terms are found because the dictionary more densely populates the space. LASSO does a better job of recovering the vectors in the sum. Much fewer than 20 vectors are usually chosen by LASSO, which is another big advantage.
\FloatBarrier
\vspace{-12pt}
\subsection{Experiment 2}
\vspace{-2pt}
This experiment was similar to the previous one, but instead of adding terms we added fact vectors from the embedded KB of the form $(-e_1+e_2)$. This is a more difficult problem for LASSO to solve because, for example, $(-e_1 + e_2) + (-e_3 + e_4)$ and $(-e_1 +e_4) + (-e_3 + e_2)$ would be exactly equal and so unrecoverable except by chance, and there are effectively twice as many entities being added. For large dictionary sizes, even two fact terms could not be reliably found. (See table 3.) 

\begin{table}
\caption{Number of relations in sum accurately recalled}
\begin{center}
\small{
\bgroup
\def\arraystretch{1.5}%  1 is the default, change whatever you need
\setlength{\tabcolsep}{0.5em} % for the horizontal padding
\begin{tabular}{|r|l|l|l|l|l|l|l|l|l|l| }
\hline
&1&	2&	3&	4&	5&	6&	7&	8&	9&	10\\
\hline
1000&	100&	100&	98&	97&	91&	90&	79&	54&	30&	5\\
10000&	100&	98&	95&	88&	85&	70&	51&	27&	7&	4\\
100000&	100&	91&	42&	30&	19&	9&	7&	4&	1&	1\\
906000&	100&	60&	25&	15&	10&	5&	1&	1&	0&	0\\
\hline
Dictionary size &
\multicolumn{10}{|l|}{}\\
\hline
\end{tabular}
\egroup
}
\end{center}
\vspace{-20pt}
\end{table}
\FloatBarrier

\subsection{Experiment 3}
\vspace{-5pt}
This experiment measured how often the system was able to find a chain of reasoning linking a given head to a tail known to be reachable in from 1 to 7 steps. We used a KB with 906000 facts, formed of all the first-order facts in CYC and conceptnet in which both entities being related could be mapped to a vector in the word2vec space (either with a corresponding English word, or as a category vector.)
\vspace{-15pt}
\begin{table}
\caption{finding paths of various lengths from a given head to a given tail}
\begin{center}
\small{
\bgroup
\def\arraystretch{1.5}%  1 is the default, change whatever you need
\setlength{\tabcolsep}{0.5em} % for the horizontal padding
\begin{tabular}{|r|l|l|l|l|l|l|l| }
\hline

&1&	2&	3&	4&	5&	6&	7\\
\hline
10000&	100&	78&	32&	33&	20&	27&	20\\
100000&	100&	92&	46&	46&	21&	31&	17\\
906000&	100&	65&	37&	35&	22&	30&	31\\
\hline
KB size &
\multicolumn{7}{|l|}{}\\
\hline
\end{tabular}
\egroup
}
\end{center}
\vspace{-25pt}
\end{table}
\FloatBarrier
\vspace{-10pt}
\section{Conclusion and Future Work}
\vspace{-5pt}
We have demonstrated how sparse decomposition methods can be used to find chains of reasoning in a knowledge graph embedded in a distributional vector space. In the future, we hope to evaluate the system on question answering datasets. The performance on longer chains needs to be improved. We would also like to find ways of integrating this method into more comprehensive cognitive architectures. The notion of antonymy in semantic vector spaces also needs a more careful treatment.

\end{document}